# Achieving detailed medial temporal lobe segmentation with upsampled isotropic training from implicit neural representation

**Yue Li[1], Pulkit Khandelwal[1], Rohit Jena[1], Long Xie[1], Michael Tran Duong[1], Amanda E. Denning[1], Christopher A. Brown[1], Laura E. M. Wisse[2], Sandhitsu R. Das[1], David A. Wolk[1] and Paul A. Yushkevich[1]**

[1]University of Pennsylvania, Philadelphia, United States
[2] Lund University, Lund, Sweden

Corresponding author: Yue Li: yue.li@pennmedicine.upenn.edu

## Abstract

Imaging biomarkers in magnetic resonance imaging (MRI) are important tools for diagnosing, tracking and treating Alzheimer's disease (AD). Neurofibrillary tau pathology in AD is closely linked to neurodegeneration and generally follows a pattern of spread in the brain, with early stages involving subregions of the medial temporal lobe (MTL). Accurate segmentation of MTL subregions is needed to extract granular biomarkers of AD progression. MTL subregions are often imaged using T2-weighted (T2w) MRI scans that are highly anisotropic due to constraints of MRI physics and image acquisition, making it difficult to reliably model MTL subregions geometrically and extract morphological measures, such as thickness. In this study, we propose a segmentation framework for MTL subregions in isotropic space, in which an implicit neural representation is used to construct the isotropic training atlas from the anisotropic low-resolution T2w data, with T1w MRI as an auxiliary modality to support the INR and segmentation. In an independent test set, the morphological measures extracted using this isotropic model showed stronger effect sizes than those from models trained on anisotropic data in distinguishing participants with mild cognitive impairment (MCI) from cognitively unimpaired individuals. In the test-retest analysis, the morphological measures extracted using the isotropic model showed greater stability than those from the anisotropic segmentation. This study demonstrates improved reliability of MRI-derived MTL subregion biomarkers without additional atlas annotation effort, which may more accurately quantify and track the relationship between AD pathology and brain atrophy for monitoring disease progression.

---

## 1 Introduction

Alzheimer's disease (AD) is the most prevalent cause of dementia. Biomarkers play a vital role in AD diagnosis, monitoring of progression, and evaluating treatment effects of disease modifying medications (Mintun et al., 2021; van Dyck Christopher H. et al., 2023). These include biomarkers from plasma, cerebrospinal fluid (CSF), and imaging (Scheltens et al., 2016). Imaging biomarkers include positron emission tomography (PET), which can target specific pathological proteins, and magnetic resonance (MR) biomarkers, which capture changes in brain structure as surrogates of downstream neurodegeneration (Jack et al., 2024).

Tau neurofibrillary tangles (NFTs) are a hallmark pathology in AD and significant pathological driver of brain atrophy. NFTs first appear in the cerebral cortex at the transentorhinal region (tERC, located on the border between entorhinal and perirhinal cortex) and gradually spread to other subregions within the medial temporal lobe (MTL), including hippocampus and amygdala, and eventually to neocortical areas in patients with AD (Braak and Braak, 1995). As a result, the MTL is the first cortical region to exhibit brain atrophy (Nelson et al., 2012). Morphological measurements of MTL subregions, including volumes of hippocampal subfields and thickness of MTL cortical subregions, serve as effective structural imaging biomarkers in magnetic resonance imaging (MRI) (Backhausen et al., 2022; Benzinger et al., 2013; Hampel et al., 2008; Márquez and Yassa, 2019).

Different MRI features influence the signal intensity and tissue contrast of MTL subfields. T2-weighted (T2w) imaging is often used to investigate the MTL because it allows for visualization of the dark band within the hippocampus (the stratum radiatum lacunosum moleculare, SRLM) (Wisse et al., 2017), which allows the boundary between CA and DG to be delineated. T1-weighted (T1w) images, however, lack such contrast within the hippocampus (Ver Hoef et al., 2021), limiting subfield segmentation. In studies and applications targeting hippocampal subfields, T2w (TSE) images at 3T field strength are often anisotropic, with an in-plane resolution of approximately 0.4 to 0.45 $mm^2$, and slice thickness ranging from 1.2 to 2.6 mm (Kreilkamp et al., 2018; Mueller et al., 2018; Romero et al., 2017; Sone et al., 2016; Wisse et al., 2016). T1w images (MPRAGE) are typically isotropic or nearly isotropic, with a resolution of approximately 0.6 to 1.2 $mm^3$, offering finer slice thickness than T2w images.

While manual segmentation is considered the gold standard for MTL subregional morphometry, automatic segmentation is essential to extract structural imaging biomarkers of MTL subregions in large datasets. Currently, mature algorithms include automatic segmentation of hippocampal subfields (ASHS) (Yushkevich et al., 2015), FreeSurfer (Fischl, 2012), and HippUnfold (DeKraker et al., 2022), all of which are atlas-based methods. The ASHS method was developed in our previous work (Yushkevich et al., 2015). It uses manually segmented hippocampal subfields (cornu ammonis 1-3 (CA1-3), dentate gyrus (DG), and subiculum (SUB)) and MTL cortex subregions (ERC, Brodmann areas 35 (BA35), Brodmann areas 36 (BA36), and parahippocampal cortex (PHC)) from T2w or T1w) images as an atlas set for multi-atlas segmentation (Yushkevich et al., 2015), with recent extensions incorporating deep learning models (Li et al., 2025b).

For FreeSurfer, the atlas was created with both ex vivo 7T 3D FLASH T2*-weighted images and in vivo T1w images, where the hippocampal subfields and surrounding brain structures were manually segmented and combined into a probabilistic model-based on a tetrahedral mesh (Iglesias et al., 2015). HippUnfold leverages the Laplacian coordinate framework to unfold the hippocampus that has been segmented from either manual or automatic processes applied to a test image. An unfolded subfield template derived from a histology image is used to assign labels to the hippocampal tissue by topologically-constrained registration (DeKraker et al., 2024, 2022).

Each of these three methods has advantages and disadvantages. The atlases for HippUnfold and FreeSurfer must be constructed from ex vivo images or histology images, which can be challenging to obtain. In contrast, the atlases for ASHS require segmentation from in vivo images, which are more readily available. Additionally, the atlases for HippUnfold and FreeSurfer focus solely on the hippocampal subfields and a small portion of extra-hippocampal regions. The atlases for ASHS contain both the hippocampal subfields and MTL cortical subregions, including the ERC, BA35, BA36, and PHC. On the other hand, the segmentation resolution achieved by HippUnfold and FreeSurfer is typically greater than the image resolution and is isotropic, while the segmentation resolution of ASHS is limited by the image resolution and is anisotropic in typical T2w images.

Previously, we developed a deep learning-based multi-modality segmentation model using the ASHS atlas to improve segmentation stability under variable image quality by making full use of the contrast and intensity advantages of different modalities (Li et al., 2025b). However, we found that the anisotropic resolution in T2w images may reduce accuracy, particularly for measuring cortical thickness. This is because the distance between structure skeleton and boundary can vary significantly depending on the direction. This raises the need for isotropic segmentation within anisotropic images.

In recent years, implicit neural representation (INR) has been widely used for fine-grained detail modeling of images at different resolutions (Mildenhall et al., 2022). This kind of methods use multi-layer perceptrons to perform continuous fitting on discrete spatial objects such as image pixels, point clouds, and meshes (Chen et al., 2021). As a continuous function, it not only offers the advantage of low storage requirements but also allows sampling of objects at arbitrary scales (Sitzmann et al., 2020), not limited by the pixel resolution of the image. In medical imaging, it achieved high-resolution sampling across multi-scale brain MRI (McGinnis et al., 2023).

In this study, we propose an isotropic MTL subregion segmentation framework. First, it takes advantage of the T1w and T2w resolutions to construct an isotropic atlas set, using a modified INR upsampling model on the existing low-resolution ASHS atlas. Then, a multi-modality segmentation model for MTL subregions is trained in this isotropic space (Li et al., 2025b). This framework was tested on an independent dataset, and morphometric measurements were extracted from these segmentations. The disease status-discrimination ability and test-retest stability analyses showed that the morphometric measurements extracted from the isotropic segmentation model can better distinguish amyloid-positive mild cognitive impairment (MCI) patients from cognitively unimpaired (CU) participants and demonstrate greater stability over time.

## 2 Methods

The proposed isotropic segmentation method consists of two training phases (Figure 1). In the first phase, we train an INR superresolution network (Section 2.2) to generate an isotropic high-resolution segmentation atlas set. In the second phase, we use this atlas to train a multi-modality segmentation model (Section 2.3). Finally, the quality of the segmentations generated by this model is corroborated using mutli-factor evaluation including assessment of test-retest stability, and assessment of ability to distinguish MCI and CU (Section 2.4). The following sections will describe these phases in detail.

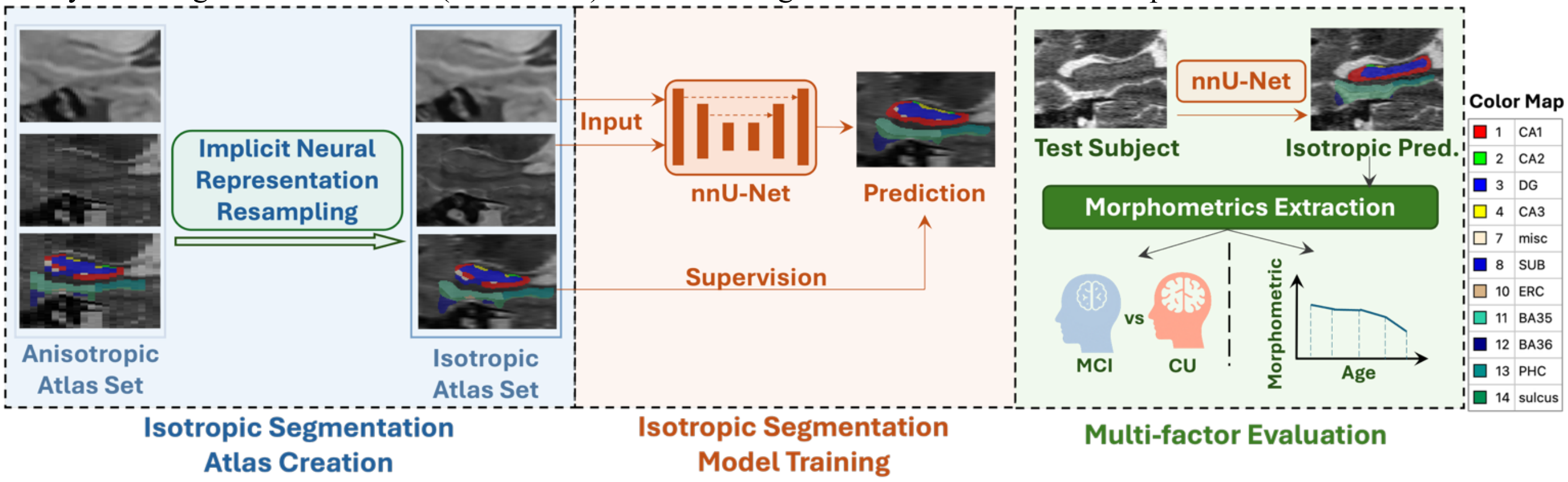

**Figure 1**. Flowchart of the isotropic subregion segmentation model. Abbreviations: MCI, mild cognitive impairment; CU, cognitively unimpaired.

### *2.1 Dataset*

#### *2.1.1 Training set*

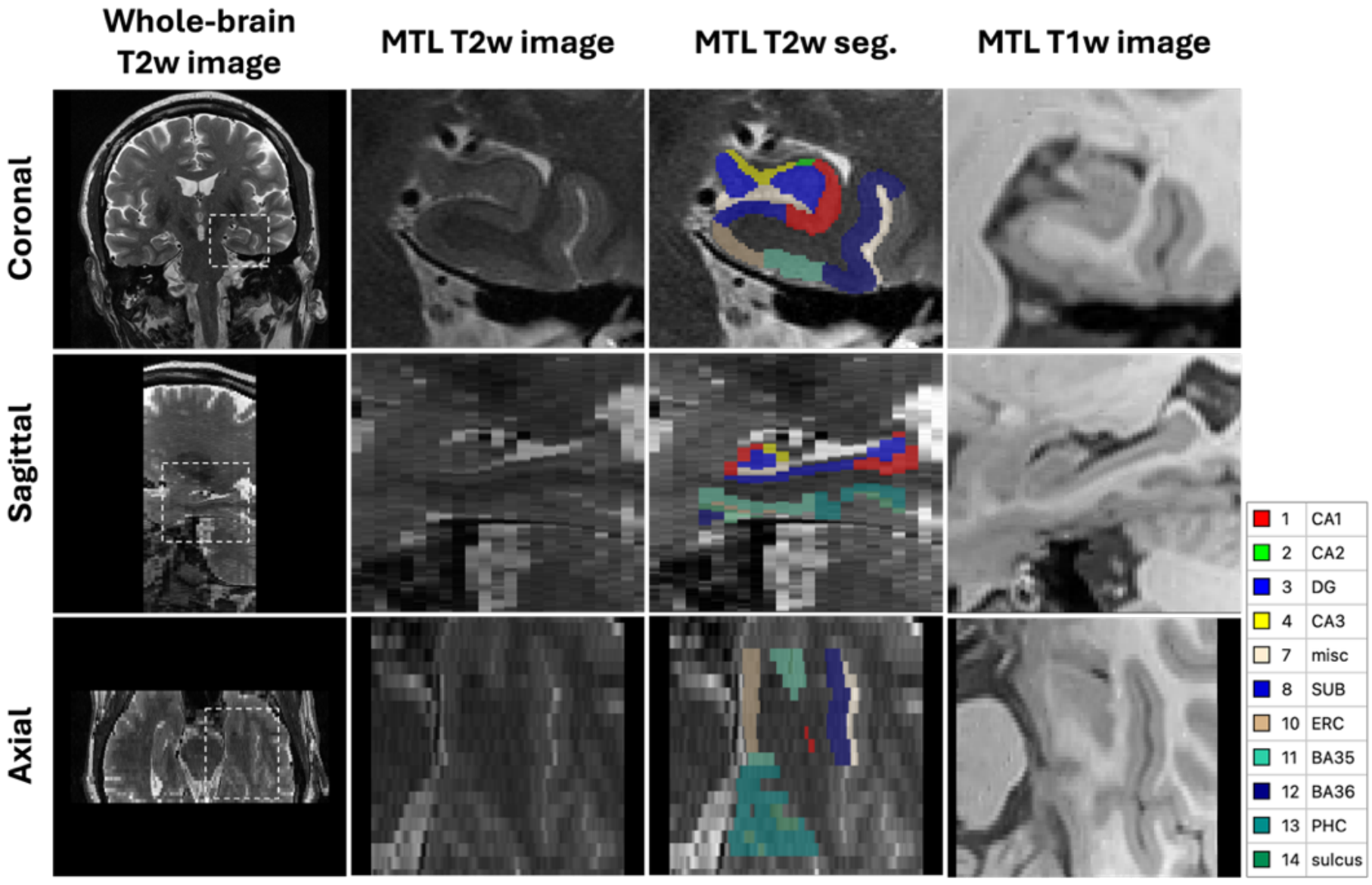

**Figure 2**. Example of images in the atlas set and corresponding manual segmentation.

An atlas set with both T1w and T2w brain MR images and manual segmentation for MTL subregions was used to train the segmentation algorithm. This atlas was originally developed in (Yushkevich et al., 2015) by manually labeling hippocampal subfields and cortical subregions, including CA1-3, DG, SUB, ERC, BA35, BA36, PHC, in the space of the T2w MRIs. It is publicly available as part of ASHS software, referred to as the T2w ASHS atlas. In later work, the atlas was updated to support MTL subregion segmentation in the space of the T1w MRI (Xie et al., 2019), referred to as the T1w ASHS atlas. The atlas contains data from 29 participants, including 15 CU participants and 14 with MCI. In addition to MTL subregions, the atlas includes a miscellaneous (MISC) category (comprising cysts, CSF in the hippocampus, and CSF near BA36) and the collateral sulcus (CS) label. In this study, the white matter (WM) surrounding the MTL was also included.

The images in the training set have different resolutions and voxel spacings. For T2w images, most have a voxel spacing of 0.4×0.4×2.6 mm³ (24 out of 29), while a smaller subset includes spacings such as 0.49×0.49×2.6 mm³ and 0.36×0.36×2.6 mm³. All T1w images are from the released T1w-ASHS dataset (Xie et al., 2019). In this dataset, the T1w images were already upsampled from an original voxel spacing of 0.98×1.0×0.98 mm³ to 0.49×0.5×0.98 mm³ (a factor-of-two in-plane upsampling) by non-local means (Manjón et al., 2010). We use this upsampled dataset directly to construct the isotropic training set in our study. An example of dataset is shown in Figure 2.

### 2.1.2 Dataset for disease status discrimination

Test set was selected from Penn Alzheimer's Disease Research Center (ADRC) cohort, where written informed consent was obtained from all Penn ADRC participants according to the Declaration of Helsinki under a protocol approved by the University of Pennsylvania Institutional Review Board.

This dataset was collected to validate whether biomarkers extracted from segmentation models can distinguish participants with MCI due to AD from the control group, referred to as status-discrimination set. According to the definition of AD, the Amyloid PET was required to provide Amyloid status. The inclusion criteria are as follows:

- The availability of both T1w and T2w MRI scans;
- A clinical diagnosis of either CU with negative amyloid status confirmed by amyloid PET (F18-Florbetaben or F18-Florbetapir) within one year of MRI or MCI with positive amyloid status confirmed by amyloid PET within one year of MRI.

This resulted in a total of 200 participants included in the test set. Demographic information for this status-discrimination set is presented in Table 1. The voxel spacings in the test set differ from those in the training set. For T2w images, among the included participants, 154 have a voxel spacing of 0.4×0.4×1.2 mm³, 35 have 0.4×0.4×1.6 mm³, 10 have 0.39×0.39×2.0 mm³, and 1 has 0.4×0.4×1.4 mm³. For T1w images, 162 have voxel spacing of 0.8×0.8×0.8 mm³ and 38 have 1.0×1.0×1.0 mm³. To maintain consistency with the training set, T1w images were preprocessed using the T1w-ASHS pipeline, resulting in upsampling in coronal slices by a factor of 2 using non-local means (Manjón et al., 2010).

**Table 1**. The demographic information of the status-discrimination test set

| | All | A- CU | A+ MCI |
|---|---|---|---|
| Number of Participants | 200 | 145 | 55 |
| Age at the MRI scanning date (years) | $71.33 \pm 5.82$ | $70.99 \pm 5.26$ | $72.24 \pm 7.07$ |
| Sex (female/male) | 129/71 | 98/47 | 31/24 |

### 2.1.3 Dataset for test-retest stability

Another dataset containing repeated MRI scans from participants over time was also collected from the Penn ADRC to validate the test-retest stability of segmentation models. The inclusion criteria for test-retest set are:

- The availability of both T1w and T2w MRI scans;
- Participant must have at least one pair of MRI scans with an interval of no more than two years between scans;
- Within the two-year period of each scan, the participant must have a diagnosis of CU status with a negative Amyloid PET.

As a result, 30 pairs of images were selected from 26 participants (22 of whom underwent two scans; 4 underwent three scans, resulting in two pairs of matched images per individual). In the total of 56 scans within the dataset, the T2w scan resolution was 0.4×0.4×1.2 mm³ for 49 scans, 0.4×0.4×1.6 mm³ for 4 scans, and 0.39×0.39×2 mm³ for 3 scans. The T1w scan resolution was 0.8×0.8×0.8 mm³ for 53 scans and 1×1×1 mm³ for 3 scans. T1w images were upsampled by a factor of 2 in coronal slice using non-local means (Manjón et al., 2010) based on T1w-ASHS pipeline.

The demographic information is shown in Table 2.

**Table 2**. The demographic information of test-retest set

| | |
|---|---|
| Number of participants | 26 |
| Number of A- CU participants* | 26 |
| Number of A+ MCI participants** | 0 |
| Number of MRI pairs | 30 |
| Sex (female/male) | 15/11 |
| Age at the first MRI scan of the pairs (years) | $71.5 \pm 5.25$ |
| Scanning-date difference of each pair (days) | $507.6 \pm 201.56$ (1.39 years) |

*A-: amyloid negative; **A+: amyloid positive

All 26 participants in the test-retest dataset were included in the status-discrimination dataset.

## 2.2 Isotropic High-resolution Segmentation Atlas Creation

In the original training atlas, T2w images exhibit higher in-plane resolution in coronal slices, whereas T1w images have higher out-of-plane resolution. To address this variable spatial resolution across the input modalities, an INR method (McGinnis et al., 2023) was employed to upsample both T1w and T2w images, along with manual segmentation into a high-resolution and isotropic space. By using a fully connected neural network, the INR can effectively learn the structure and intensity features of the MTL by establishing an implicit mapping between image coordinates and voxel values, respectively. The following outlines the pipeline for creating the high-resolution atlas.

### 2.2.1 Preprocessing

This study focused on segmenting MTL subregions. The MTL regions of interest (ROIs) for T2w and T1w were obtained from the original lower resolution T2w ASHS atlas set and T1w ASHS atlas set, respectively. These two atlas sets were constructed using images from the same cohort of participants, enabling us to utilize paired T2w and T1w ROIs. For the hippocampus, we aim to segment its subfields. However, these labels are only available in the T2w atlas, while the T1w atlas set contains only labels for the anterior and posterior hippocampus. Therefore, in subsequent steps, we retained only the manual segmentation from the T2w atlas.

Since INR does not perform image alignment, it is necessary to register different imaging modalities to one another for the same participant. We performed rigid alignment to register the T1w ROI to the T2w ROI using the Greedy registration tool (Venet et al., 2021). By default, Greedy reslices the aligned image to the fixed image's resolution. In our case, however, we do not resample the moving image to the fixed image resolution. Instead, the alignment is accomplished by writing the rigid transform into the header of the T1w image, so that each modality keeps its own voxel grid and resolution. To ensure that the bounding box of patches in physical space align with one another, the image array of the T1w patch was cropped or padded to fit the same physical spatial bounding box as the T2w ROI.

### 2.2.2 Implicit Neural Representation Network Architecture

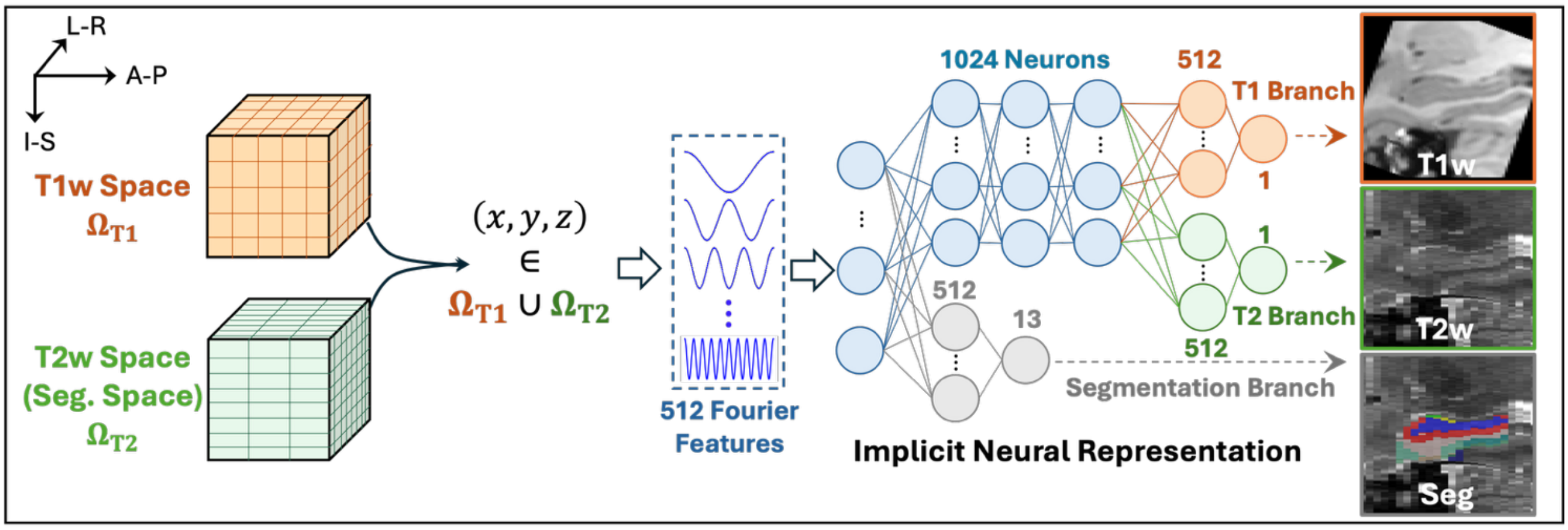

**Figure 3**. The architecture of implicit neural representation to establish isotropic segmentation atlas.

The key idea of the INR here is to upsample the manual segmentation from anisotropic to isotropic resolution, so that we can build the isotropic training set for MTL subregion segmentation. A well-trained INR turns the discrete image into a continuous representation of the MTL structure. During its training, the manual segmentation, the T2w image, and the T1w image each play a different role. The segmentation gives the anatomical prior for the subregion boundaries; the T2w image provides the intensity information and the spatial connection with the manual segmentation, which was annotated in the T2w space; and the T1w image contributes its higher resolution along the slice thickness. By making full use of the strength of each modality, the in-plane detail of the T2w image and the slice-thickness detail of the T1w image, the continuous representation captures the MTL structure well, and the isotropic segmentation can then be sampled from it.

Figure 3 shows the network architecture of INR, adapted from (McGinnis et al., 2023). The image array coordinates of aligned T1w and T2w ROIs were normalized into a standard scale with the top-left corner having a coordinate of (-1, -1, -1) and the bottom-right corner having a coordinate of (1, 1, 1). All voxels in both ROIs were linearly assigned new coordinates with values between -1 and 1 according to their grid locations in the ROIs.

A column vector $\boldsymbol{v}$ of 512 Fourier features (Tancik et al., 2020) was extracted from each three-element coordinate $\boldsymbol{x}$ using the formula $\boldsymbol{v} = [\cos(2\pi\boldsymbol{x}\boldsymbol{b}^{\boldsymbol{T}})\,;\ \sin(2\pi\boldsymbol{x}\boldsymbol{b}^{\boldsymbol{T}})]^T$, where $\boldsymbol{x} = (x, y, z)$ represents the normalized coordinate and $\boldsymbol{b}$ is a matrix with size of $3 \times 256$ sampling from a Gaussian distribution with mean value of 0 and standard deviation of 4. These Fourier features were fed into a fully-connected (FC) network. The network contains four shared layers (blue layers in Figure 3), with each including an FC module, and a rectified linear unit (ReLU) module. The first shared layer has input size of 512, which represents the size of Fourier features, and output size of 1024. All other shared layers have both an input size and output size of 1024, indicating the hidden feature size. These hyperparameters are from the default setting in study (McGinnis et al., 2023).

Then, another three task-specific FC subnets are attached after these shared layers, separately. Each subnet contains two FC layers with ReLU activation in-between. These subnets are responsible for predicting the voxel values in T1w ROI, T2w ROI, and the manual segmentation in T2 ROI, respectively. The T1w branch and T2w branch each output one value, representing the intensity of T1w and T2w image at the input coordinate. The segmentation branch is connected after the first shared layer instead of the last one. This prevents segmentation supervision during training from overly interfering with the other two branches, thereby avoiding the generation of fake border between subregions in the resulting T1w and T2w images (examples are available in supplementary materials S1). This branch outputs 13 values that represent the predicted one-hot segmentation.

The ground truth for training this network consists of the corresponding voxel values of coordinate samples, including intensity values for T1w or T2w ROIs, as well as one-hot encoded labels for the segmentation ROI. The intensity of each image is normalized to a range of 0 to 1 by the Min-Max scaler.

The supervision of the INR training is the reconstruction at the observation points. If the training is regarded as a non-linear interpolation, each grid coordinate of each modality is an observation point. The goal of training is therefore to make sure that the value at each of these observation points can be reconstructed. To this end, the mean squared error loss was used to supervise the voxel intensity prediction of the T1w and T2w images, and the binary cross-entropy loss was used to supervise the segmentation prediction. The final loss function is shown below:

$$L_{Final} = \sum_{x\in\Omega_1} (\widehat{I_1}(x) - I_1(x))^2 + \sum_{x\in\Omega_2} (\widehat{I_2}(x) - I_2(x))^2 - \sum_{x\in\Omega_2}\sum_{i=1}^{13}[y_i(x)\cdot\log\widehat{y_i}(x) + (1 - y_i(x))\log(1 - \widehat{y_i}(x))]\,,$$

where $\Omega_1$ is the coordinate space of the T1w ROI, $\Omega_2$ is the coordinate space of the T2w ROI, $I_1$ and $I_2$ are the T1w and T2w arrays, number 13 is the number of labels in the segmentation ROI, $y_i$ is the ground-truth of label $i$, and all letters with a hat are the corresponding predicted results.

Adam optimizer was employed as the optimizer for INR parameter optimization. The initial learning rate was set to 0.0004 and was adjusted during training using Cosine Annealing.

### *2.2.3 Isotropic Hyper-resolution Atlas*

An INR model was trained for each ROI (left and right sides) of each participant in the atlas set. Each model was trained for 60 epochs. Once trained, each model is a function $f(\boldsymbol{x})$ that characterizes the intensity and segmentation of an individual MTL structure in continuous space. To obtain a discrete upsampled image, we evaluate this function using a uniform sampling grid with resolution of $0.4\times0.4\times0.4$ mm$^3$. By extracting the Fourier features of these coordinates and inputting them into each trained INR for inference, we obtain synthetic upsampled T1w and T2w images and corresponding segmentation images with isotropic voxels. The INR-upsampled MTL segmentations are then used to train the segmentation model in the second phase of our method.

## 2.3 Multi-modality Segmentation Model

### 2.3.1 Model Training

The segmentation model (phase two of our method) uses the nnU-Net (Isensee et al., 2021) framework, with additional modality augmentation (Li et al., 2025b) to allow for missing or corrupted modalities. Specifically, when training the segmentation model, the T1w or T2w input image is replaced with Gaussian noise at a probability of 50%. This forces the network to extract information from both modalities, rather than focusing on one and neglecting the other, so that the model can handle corrupted quality or a missing modality.

We trained the model using INR-upsampled segmentations output from phase one. However, the INR-upsampled T1w and T2w images from phase one were not used for segmentation model training, because INR upsampling is time-consuming and requires at least 30 minutes on a NVIDIA Quadro RTX 5000 GPU to train for 60 epochs, which is impractical when performing inference with the segmentation model. Instead, at inference, the segmentation model would be given isotropic T1w and T2w images upsampled using fast trilinear interpolation. To minimize domain shift between training and inference for the segmentation model, we trained the segmentation model using linearly upsampled T1w and T2w scans, and INR-upsampled segmentations.

During training, the model took the T2w image as primary modality, and the T1w image was registered to the T2w image rigidly again, but in isotropic space, to correct possible subtle misalignment. Soft Dice loss and cross-entropy loss were used jointly to supervise the training of the segmentation model and a modality augmentation (Xie et al., 2023) scheme was used during training to avoid overdependence of the model on the primary modality.

### 2.3.2 Model Inference

The test set consists of whole-brain T1w images and T2w dedicated images targeting in the hippocampal region. The ROIs containing the left and right MTL were to be extracted as follows.

- We used a T1w template (available in the ASHS package) that includes the left and right MTL ROI masks, with the left ROI measuring 49.8×56.0×78.1 $mm^3$ (left-right×anterior-posterior×superior-inferior) and the right ROI measuring 48.8×63.0×90.8 $mm^3$.
- During ROI extraction, the neck region in the T1w image was first trimmed and the whole image was rigidly aligned with the T2w image.
- Then, the aligned T1w image was sequentially registered to the template using rigid, affine, and deformable registrations. This process was implemented by the Greedy tool (Venet et al., 2021) using the same parameters in original ASHS pipeline (Yushkevich et al., 2015).
- Finally, the inverse deformation field from the above registrations was applied to transform the left and right template ROI masks from the template space into the individual space.
- The minimum bounding box containing the left or right mask was used to crop the T1w and T2w MTL ROIs from the corresponding participant.

After obtaining the T1w and T2w ROIs, similar to the process during training, they were first linearly upsampled to an isotropic space with a resolution of $0.4 \times 0.4 \times 0.4\ \mathrm{mm}^3$. Subsequently, the T1w ROI was rigidly registered to the T2w ROI, forming the two-channel input fed into the trained model to finish segmentation inference.

## 2.4 Evaluation

### 2.4.1 Baseline Models

Two models were employed as baselines in this study:

- *ASHS* (negative control): Our research framework was constructed based on ASHS, which is currently the most mature MTL cortical subregion segmentation method in our study. Therefore, the classic ASHS (Yushkevich et al., 2015) based on multi-atlas segmentation was used as a baseline for comparison with the method proposed in this paper.
- *SuperRes-Seg* (positive control): A super-resolution model is trained on paired in vivo 3T T2w images and 9.4T post-mortem MRI. It upsamples the manual segmentation in the original 3T T2w space by a factor of five along the slice thickness. This method can upsample only the segmentation, not the T1w and T2w images, because the intensity distribution and the detailed contrast differ between the 3T T2w and 9.4T MRI. In this study, the segmentation upsampled by this method and the images upsampled by linear interpolation (the same image-upsampling method as in the proposed model, for a fair comparison) are used to train the multi-modality nnU-Net, which we refer to as *SuperRes-Seg* method. We note that, since the upsampling factor of this method is fixed at five, the final resolution is a nearly isotropic 0.4×0.4×0.52 mm³, where 0.52 is one fifth of 2.6, the most common slice thickness in the original atlas.

- *Linear-Seg* (positive control): Since SuperRes-Seg can only reach a nearly isotropic resolution, we wanted to include another baseline that can provide a real isotropic resolution for the segmentation. Given the lack of a paired anisotropic- and isotropic-resolution training dataset, an intuitive approach is to upsample the atlas (both images and segmentations) to an isotropic resolution using linear interpolation, and then train a multi-modality segmentation model on this data. We refer to this as the Linear-Seg method and use it as another baseline. Specifically, for manual segmentations, trilinear upsampling to 0.4×0.4×0.4 mm³ resolution was performed after one-hot encoding. Direct upsampling can change the resolution but cannot reduce the stair-like effect on the boundary. Therefore, a Gaussian kernel with a standard deviation of 0.75 voxels was applied on each binary map in one-hot encoding to smooth segmentation. Finally, argmax voting is used to obtain a multi-label isotropic segmentation. The choice of the standard deviation parameter was derived from observations of the sample data and specific examples are provided in the supplementary materials S2.

The default nnU-Net on anisotropic space was not used as one baseline in this study. This was because its preprocessing step resizes test images to a fixed resolution that determined from training set. Since the test set resolution is higher than that of training set in this study, resizing the test set would reduce the segmentation resolution, and further result in adjacent slices receiving identical segmentation outputs, which is inaccurate. Conversely, omitting this resizing step would constrain the network's receptive field during inference, leading to wrong segmentation. Examples illustrating both scenarios are provided in the supplementary materials S3.

Since FreeSurfer and HippUnfold are also segmentation methods for hippocampal subfields, we ran them on our test set and evaluated them on the same downstream tasks. However, there are significant differences in subregion definitions, coverage scope, segmentation protocols, and specific segmentation approaches between ASHS and these methods. We did not conduct direct comparisons between them but present these results in the supplementary materials S4 and S5 for reference.

### 2.4.2 Task-based Evaluation

Because manual annotation is time-consuming, a large external test set with manual segmentations following the protocol used in our training set was not available to test the generalization ability of our segmentation model. Therefore, the Dice coefficient was calculated only on the atlas using five-fold cross-validation. Then, the average Dice score for each subregion or subfield was computed. For both the linear upsampling model and our proposed INR-based upsampling model, the Dice was calculated not only in the high-resolution isotropic space, but also in the original anisotropic space after resampling the model's validation output back to space of original manual segmentation.

Considering the final goal of MTL subregion segmentation is to extract imaging biomarkers to diagnose and monitor AD progression, several downstream tasks correlated to biomarker quality were used to evaluate the model's performance in test sets.

Measures of interest extracted from MTL segmentations included volumes of hippocampal subfields and median thickness of MTL cortical subregions (ERC, BA35, BA36, PHC). The thickness calculation was performed using the open-source tool *cm-rep* (Pouch et al., 2015), which extracted the pruned Voronoi skeleton (Ogniewicz and Kübler, 1995) and computed the median value of the radius field over the portion of the skeleton belonging to each subregion.

The following downstream evaluation tasks were then performed:

- *Ability to distinguish CU and MCI.* After calculating the measures of interest in the status-discrimination test set, we used a general linear model (GLM) to test whether each measure significantly differed between CU and MCI groups. Intracranial volume (ICV), age and sex were used as covariates for volume analyses, and age and sex was used as covariate for thickness analyses. In addition to analyzing single measure, we also fitted a single multivariate GLM using all volume measures of hippocampal subfields, with ICV and age as covariates, and a single multivariate GLM using all thickness measures of corical subregions, with age as covariate. P-values and area under the receiver operating characteristic curve (AUC) values were used as evaluation metrics to perform comparison.
- *Test-retest consistency in CU participants*. For CU participants in test-retest set, biological changes in hippocampal subfield volume and cortical thickness within two years are expected to be small (0.84% for one year and 1.76% for two years for hippocampal volume according to the research (Fjell et al., 2009)). Therefore, scans from different time points for the same participant collected within two years should show relatively consistent biomarkers measurements. To measure the test-retest stability of biomarkers, for volume, we calculate the percentage difference between the two scans' volumes relative to their average volume; for median thickness, we calculate only the difference in median thickness between the two scans. Then, two metrics were used to measure the consistency of each pair, the standard deviation of change over time and total absolute change over time in the whole test-retest set. The specific formula can be found in the supplementary materials S6.

- *Pointwise comparison between CU and MCI in template space*. In addition to the median thickness, the pointwise thicknesses of CA, SUB and cortical subregions were also used to differentiate between CU group and MCI group after all segmentation results were modeled using a fitted surface-based representation and mapped to an MTL surface template. The pipeline CRASHS (Yushkevich et al., 2024) calculated thickness by finding the mid-surface between the inner and outer surfaces of the cerebral cortex and constructing a mapping between individuals and template in terms of this mid-surface. The *meshglm* tool evaluated group differences with visualizations by t-statistic maps.

### 2.5 Computing Environment

The deep learning models were trained and tested using PyTorch 2.5.1 on a GPU cluster with the Ubuntu 24.04.3 operating system and CUDA 12.2. The CPU used is an AMD Ryzen Threadripper 3970X 32-Core Processor, while the GPUs are NVIDIA Quadro RTX 5000.

## 3 Results

### 3.1 Visualization of Upsampled Results

The INR was trained for each sample. By design, we did not have an upsampled ground truth to supervise it. Therefore, we could only obtain the results of sampling according to the INR after the training loss converged. Figure 4 shows illustrative examples of INR upsampling of the manual segmentation in the atlas set. For comparison, it also shows the original segmentation at the anisotropic resolution (used to train the baseline ASHS), the super-resolution result at the nearly isotropic resolution of 0.4×0.4×0.52 mm³ (used to train SuperRes-Seg), and the linearly upsampled result at the isotropic resolution of 0.4×0.4×0.4 mm³ (used to train Linear-Seg).

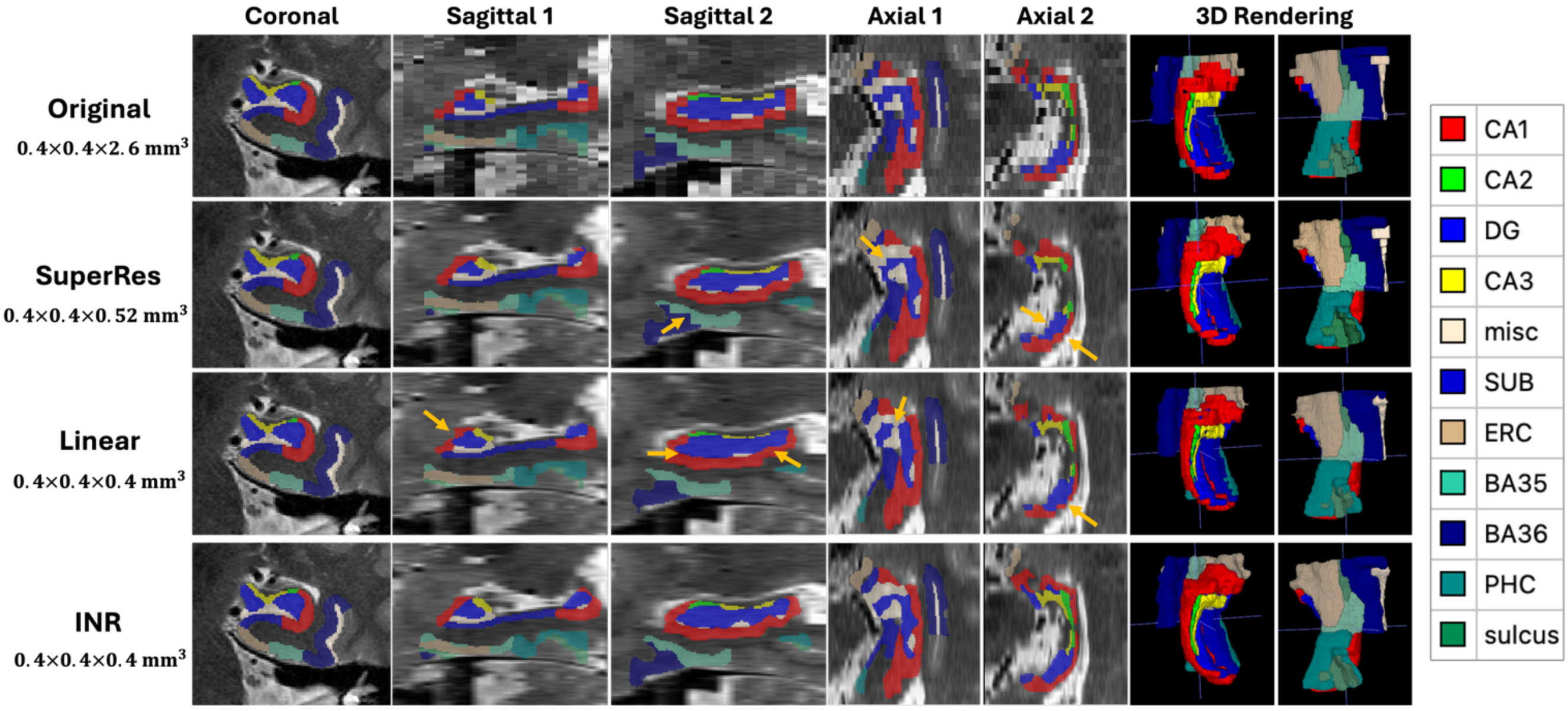

**Figure 4**. Examples of manual segmentation at original resolution, nearly isotropic resolution upsampled by SuperRes method, isotropic resolution upsampled by linear method and INR method. The orange arrows point out the locations where these two upsampling methods showed obvious problems.

As shown in Figure 4, these upsampling methods kept the center coronal slice unchanged. In the sagittal and axial views, the blocky boundary in the anisotropic space was smoothed after upsampling by these methods. However, SuperRes upsampling still did not smooth the boundary everywhere. For example, the borders between BA35 and BA36, between DG and misc, and between CA1 and the background (pointed out by orange arrows) still showed a stair-like shape. The same artifact remained visible in the 3D rendering, even though the surface was much smoother overall.

The Linear upsampling method showed a different kind of problem. It tended to lose thin or small regions. In the sagittal 1 and axial 1 and 2 examples, the boundary regions were under-segmented and part of the structure was eroded (orange arrows). In the sagittal 2 example, the misc region disappeared entirely (orange arrows).

INR method smoothed the boundary from the blocky anisotropic space while keeping the detailed subregions from being lost. It produced a smooth boundary with no remaining stair-like shape. It also preserved the small regions, so that the detailed structure present in the anisotropic space still existed in the INR-upsampled space. In both respects, it outperformed the other two upsampling methods in the qualitative comparison.

### 3.2 Cross validation in training set

Using the same cross-validation splits (five folds), three baseline methods, including ASHS, SuperRes-Seg method and Linear-Seg method, along with our proposed method, were trained on the atlas dataset. For each subregion, we computed its average Dice coefficient between ground-truth and automatic segmentations by cross-validation. For SuperRes-Seg, Linear-Seg, and our proposed method, the ground truths are at different resolutions, so a direct comparison of their Dice scores is not fair. We therefore computed the Dice coefficient not only in each method's training space, but also in the original anisotropic space, using the raw manual segmentation as the reference. For this, we resampled the predicted segmentation onto the anisotropic grid with the greedy toolbox, using an identity transform and its label reslicing mode, so that only the sampling grid changed and no spatial alignment was applied. This enabled direct comparison across all three models. The average Dice scores of validation for all subregions and subfields are shown in Table 3. The specific p-values of the Wilcoxon signed-rank test for the comparison are in Supplementary Material S8.

**Table 3**. Average Dice scores in cross-validation. For each subregion in the anisotropic space, the best result is shown in bold blue, and the second-best result is underlined. (The p-values of the Wilcoxon signed-rank test between each baseline and the proposed method in the anisotropic space are marked with stars: * $p < 0.05$, ** $p < 0.01$, *** $p < 0.001$.)

| **Model** | **Anisotropic Model** | **SuperRes-Seg Method** | | **Linear-Seg Method** | | **Proposed Model** | |
|---|---|---|---|---|---|---|---|
| **Evaluation Space** | **Anisotropic** | **Nearly Isotropic** | **Downsampled to Anisotropic** | **Isotropic** | **Downsampled to Anisotropic** | **Isotropic** | **Downsampled to Anisotropic** |
| **CA1** | 0.804 *** | 0.813 | 0.802 *** | 0.826 | <u>0.812</u> *** | 0.805 | **0.829** |
| **CA2** | 0.557 *** | 0.614 | 0.540 *** | 0.572 | <u>0.602</u> *** | 0.625 | **0.649** |
| **CA3** | 0.535 *** | 0.562 | <u>0.567</u> *** | 0.514 | 0.519 *** | 0.571 | **0.601** |
| **DG** | 0.823 *** | 0.810 | 0.802 *** | 0.862 | <u>0.843</u> *** | 0.833 | **0.852** |
| **SUB** | 0.750 *** | 0.776 | 0.755 *** | 0.789 | <u>0.774</u> *** | 0.770 | **0.786** |
| **ERC** | 0.777 *** | 0.810 | 0.799 *** | 0.827 | <u>0.806</u> *** | 0.798 | **0.817** |
| **BA35** | 0.702 *** | 0.734 | 0.740 *** | 0.768 | <u>0.755</u> ** | 0.743 | **0.764** |
| **BA36** | 0.771 *** | 0.814 | 0.806 *** | 0.828 | <u>0.808</u> *** | 0.804 | **0.822** |
| **PHC** | 0.782 *** | 0.816 | 0.808 *** | 0.846 | <u>0.828</u> * | 0.814 | **0.834** |

If we focus only on the Dice scores relative to the ground-truth (i.e., in the original anisotropic space), our proposed INR-based model achieved the highest Dice score in all subregions. This demonstrates the improvement in segmentation accuracy achieved by the INR-based upsampling scheme.

When comparing the evaluations in the isotropic and anisotropic spaces within each model, the Dice scores of the SuperRes-Seg and Linear-Seg methods decreased after the predicted segmentation was downsampled to the raw anisotropic space, while the Dice score of our proposed INR-based method improved. A possible explanation is that the interpolation or upsampling used to generate the ground truth may not fully reflect the real anatomical structure. When the model learns from such a ground truth, its prediction can deviate from the true annotation, so the Dice decreases once it is compared against the raw manual segmentation in the anisotropic space. For our proposed method, the Dice increased instead, which may indicate that the INR upsampling produces a ground truth closer to the real anatomical structure.

### 3.3 Visualization of predicted segmentation

After models were trained, we evaluated our proposed INR-based segmentation model and the other three baseline models on two test datasets. We randomly selected five examples from the test sets and present them in Figure 5. For example 1, we display two slices from sagittal view and a 3D rendering. For the other examples, only the 3D rendering is shown.

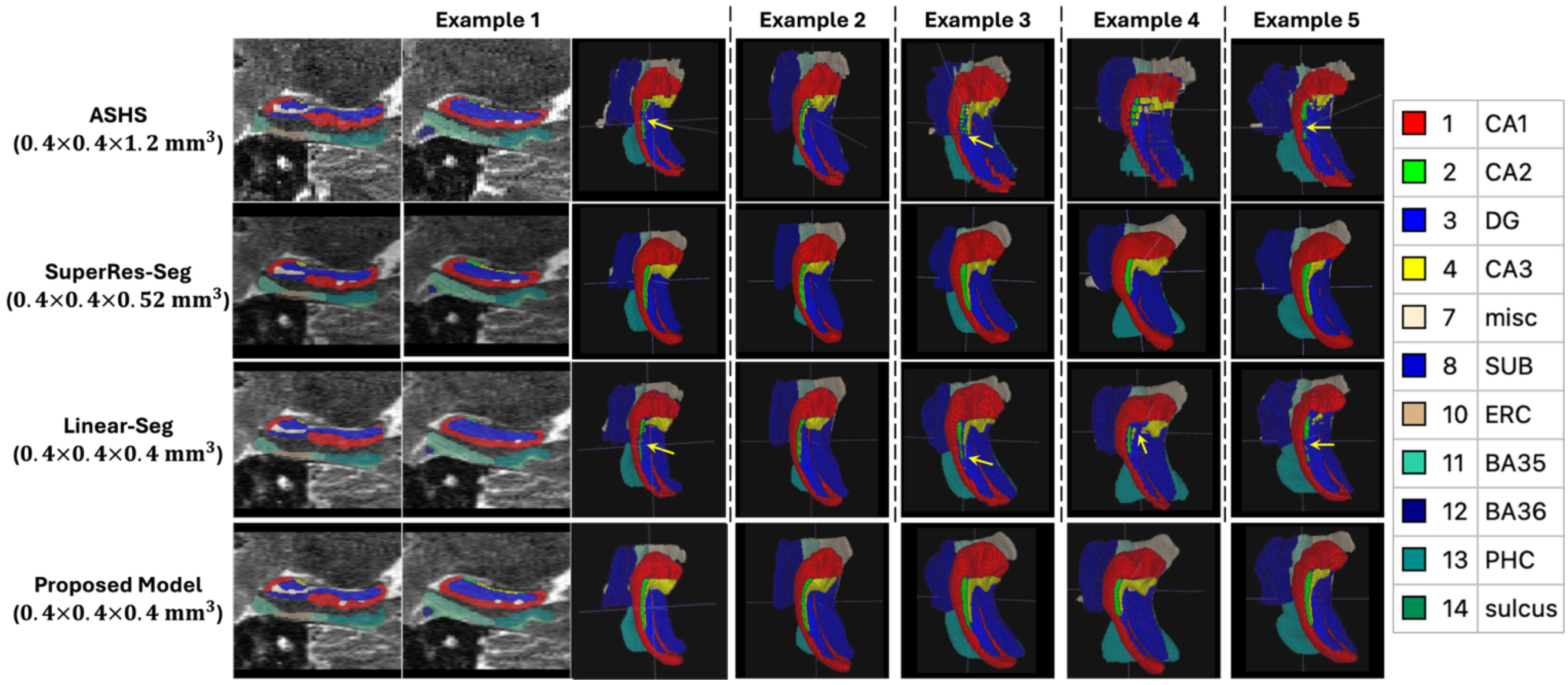

**Figure 5.** Segmentation examples from different models in the test set.

Comparing these models, the segmentations by ASHS show an obvious blocky boundary in the 3D rendering views, since they were obtained in the anisotropic space. For thin subfields like CA2 and CA3, the ASHS segmentations also show under-segmentation in certain cases, as pointed out by the yellow arrows. The Linear-Seg method performed better on surface smoothness, but it still under-segmented CA2 and CA3. The SuperRes-Seg and the proposed model generated similar results in these cases, both in smoothness and in the segmentation of the thin subfields, where no under-segmentation was found.

Although the SuperRes method upsampled the segmentation with a relatively blocky boundary in the training set, the inference by SuperRes-Seg in the test set looks smooth. The smooth test result could benefit from the strong adaptation ability of the nnU-Net, which is not affected by the noisy boundary. We should emphasize, however, that the SuperRes-Seg runs in the nearly isotropic space because it is limited by the training resolution, and this is one of the limitations of this method.

### *3.4 Ability to distinguish CU vs MCI*

In the status-discrimination dataset, we calculated the volumes of hippocampal subfields and the thicknesses of cortical subregions from the model segmentation results as measures of interest. We then fitted GLMs to individual measures to evaluate their ability to differentiate A-CU and A+MCI groups and multivariate GLMs were also fitted by considering all subregions' volumes or thicknesses together in the same side.

In 10 participants, the calculation of subregion thickness in ASHS segmentation was failed because the extracted skeleton extended beyond the subregion boundaries, resulting in NaN being introduced into the labels. Our proposed method and other two baseline methods did not encounter these failures. For fairness of comparison, all GLM models here were conducted in the remaining 190 participants only. P-values for the regression coefficient of the GLM fitting and AUC values are listed in DeLong's test was used to compare the AUCs between each baseline and the proposed model. The proposed model's AUC was significantly higher than that of ASHS for left CA1, left CA2, right CA2, left PHC, and right PHC ($p < 0.05$). The proposed model also significantly outperformed SuperRes-Seg for right CA1 and left BA35 ($p < 0.05$), while SuperRes-Seg significantly outperformed the proposed model ($p < 0.05$) on both the left and right DG.

**Table** *4* (for hippocampal subfield volume) and

Table **5** (for cortical subregion thickness).

It can be observed that no model achieved the highest AUC values or the smallest p-values across all subregions in the single-variate GLM evaluation.

In the multi-variate GLM, the proposed model achieved the highest AUC values, and the Linear-Seg method achieved the second-best AUC values, for the left hippocampal subfield volume, the right hippocampal subfield volume, and the left cortical median thickness. For the right cortical median thickness, the best and second-best places were swapped between these two models. The ASHS model, which segments the subregions in the anisotropic space, had the lowest AUC values in most subregions, in both the single-variate and multi-variate GLM evaluations.

Comparing the same subregions or multi-variate GLMs between the left and right sides, we found that the left-side measurements distinguished CU from MCI better than the right-side measurements. This is consistent with the common understanding of AD, in which the left side shows atrophy earlier than the right side.

DeLong's test was used to compare the AUCs between each baseline and the proposed model. The proposed model's AUC was significantly higher than that of ASHS for left CA1, left CA2, right CA2, left PHC, and right PHC ($p < 0.05$). The proposed model also significantly outperformed SuperRes-Seg for right CA1 and left BA35 ($p < 0.05$), while SuperRes-Seg significantly outperformed the proposed model ($p < 0.05$) on both the left and right DG.

**Table 4**. General linear model to evaluate hippocampal subfield volumes' ability to distinguish cognitively unimpaired and mild cognitive impairment (with age, sex and intracranial volume as a covariate). P-values in the table indicate the significance of the group effect on each measurement, adjusted for age, intracranial volume and sex.The best result in each row is highlighted in blue and the second-best result is underlined. Asterisks denote DeLong's test comparing the AUCs between each baseline and the proposed method (* p_DeLong < 0.05, ** p_DeLong < 0.01, *** p_DeLong < 0.001).

| Side | Subregion | ASHS (Anisotropic) | | SuperRes-Seg (Nearly isotropic) | | Linear-Seg (Isotropic) | | Proposed (Isotropic) | |
|---|---|---|---|---|---|---|---|---|---|
| | | *p*-value ↓ | AUC ↑ | *p*-value ↓ | AUC ↑ | *p*-value ↓ | AUC ↑ | *p*-value ↓ | AUC ↑ |
| Left | CA1 | 2.05E-16 | 0.848* | 3.00E-17 | <u>0.858</u> | <u>1.32E-17</u> | 0.855 | **9.50E-19** | **0.87** |
| | CA2 | 3.00E-05 | 0.71** | **3.99E-13** | **0.82** | 3.74E-11 | 0.788 | <u>7.46E-13</u> | <u>0.809</u> |
| | CA3 | 0.621 | 0.483 | 0.697 | 0.515 | 0.888 | 0.486 | 0.769 | 0.497 |
| | DG | 5.07E-11 | 0.792 | **7.08E-16** | **0.848**** | 1.22E-13 | <u>0.823</u> | <u>1.08E-13</u> | 0.821 |
| | SUB | <u>6.18E-09</u> | <u>0.767</u> | 1.09E-08 | 0.759 | **3.43E-09** | **0.768*** | 3.06E-08 | 0.749 |
| | All | | 0.875 | | 0.885 | | <u>0.887</u> | | **0.895** |
| Right | CA1 | 6.94E-14 | 0.815 | <u>2.07E-15</u> | 0.818* | 3.00E-15 | <u>0.828</u> | **2.47E-16** | **0.833** |
| | CA2 | 2.32E-06 | 0.736** | <u>3.65E-15</u> | 0.829 | **2.85E-15** | **0.837** | 8.68E-15 | <u>0.835</u> |
| | CA3 | 0.326 | 0.526 | 0.39 | 0.548 | 0.218 | 0.553 | 0.317 | 0.535 |
| | DG | 1.67E-09 | 0.764 | **4.53E-13** | **0.808**** | <u>1.31E-12</u> | <u>0.795</u> | 2.28E-11 | 0.781 |
| | SUB | **6.57E-09** | **0.766** | 1.06E-06 | 0.739 | <u>1.14E-07</u> | <u>0.752</u> | 1.93E-07 | 0.749 |
| | All | | 0.83 | | 0.857 | | <u>0.864</u> | | **0.874** |

**Table 5**. General linear model to evaluate cotical thickness's ability to distinguish cognitively unimpaired and mild cognitive impairment (with age and sex as a covariate). P-values in the table indicate the significance of the group effect on each measurement, adjusted for age and sex. The best result in each row is highlighted in blue and the second-best result is underlined. Asterisks denote DeLong's test comparing the AUCs between each baseline and the proposed method (* p_DeLong < 0.05, ** p_DeLong < 0.01, *** p_DeLong < 0.001).

| Side | Subfield | ASHS (Anisotropic) | | SuperRes-Seg (Nearly isotropic) | | Linear-Seg (Isotropic) | | Proposed (Isotropic) | |
|---|---|---|---|---|---|---|---|---|---|
| | | *p*-value ↓ | AUC ↑ | *p*-value ↓ | AUC ↑ | *p*-value ↓ | AUC ↑ | *p*-value ↓ | AUC ↑ |
| Left | ERC | 1.50E-05 | 0.694 | 1.10E-06 | 0.72 | <u>2.80E-07</u> | **0.724** | **2.04E-07** | <u>0.722</u> |
| | BA35 | <u>0.000462</u> | <u>0.668</u> | 0.0166 | 0.615* | 0.00139 | 0.651 | **0.000319** | **0.671** |
| | BA36 | <u>6.58E-05</u> | 0.679 | 0.0053 | 0.652 | 0.000198 | <u>0.68</u> | **4.44E-05** | **0.686** |
| | PHC | 0.3 | 0.521* | 0.0362 | 0.595 | **0.0178** | **0.611** | <u>0.023</u> | <u>0.602</u> |
| | All | | 0.73 | | 0.73 | | 0.748 | | **0.758** |
| Right | ERC | 0.00419 | 0.634 | 2.98E-06 | 0.687 | **5.90E-07** | **0.712** | <u>1.34E-06</u> | <u>0.7</u> |
| | BA35 | **0.00472** | 0.615 | 0.0328 | 0.605 | 0.0171 | <u>0.623</u> | <u>0.00575</u> | **0.629** |
| | BA36 | **0.00233** | **0.643** | 0.0508 | 0.59 | 0.00349 | <u>0.641</u> | <u>0.00281</u> | 0.622 |
| | PHC | 0.648 | 0.508*** | **0.0078** | **0.628** | 0.0272 | 0.581 | <u>0.00988</u> | <u>0.606</u> |
| | All | | 0.686 | | 0.688 | | **0.722** | | <u>0.701</u> |

The results obtained by HippUnfold and FreeSurfer on the same dataset are available in the supplementary materials S4 for reference. In the multivariate model, HippUnfold's AUC values were lower than those of our method, while FreeSurfer's AUC values were slightly higher. However, these methods differ in label definitions, protocols, training data, and other aspects, so we did not perform a direct comparison here.

### *3.5 Test-retest consistency in CU subjects*

In the test-retest set composed of CU participants, the temporal stability of subfield volume and subregion thickness was analyzed. Specifically, as defined in Section 2.4.2, for the volume of subfields, the difference over time was expressed as percentage; whereas for subregion thickness differences, the absolute difference was directly used. It should be noted that during thickness calculations, ASHS failed in 5 paired images across 4 participants due to skeleton extending beyond the subregion. Therefore, for fairness, thickness comparisons were conducted using the remaining 25 scan pairs from the 22 participants. The specific comparisons are shown in Table 6 and Table 7.

In the volume analysis, the SuperRes-Seg method achieved the smallest SD in most subfields, especially on the left side. Our proposed method achieved the second-best result in most subfields and the best result in a few of them. In the thickness analysis, our proposed method achieved the smallest SD in most subregions, while the second-smallest SD was obtained by the SuperRes-Seg method in most subregions. In general, these nearly isotropic and isotropic models outperformed the anisotropic model, ASHS, in most subregions.

**Table 6.** Subfield volume test-retest stability analysis. The proposed INR-based method achieved better stability. The best result in each row is highlighted in blue and the second-best result is underlined.

| Side | Subregion | ASHS (Anisotropic) | | SuperRes-Seg (Nearly isotropic) | | Linear-Seg (Isotropic) | | Proposed (Isotropic) | |
|---|---|---|---|---|---|---|---|---|---|
| | | SD ↓ | AbsSum ↓ | SD ↓ | AbsSum ↓ | SD ↓ | AbsSum ↓ | SD ↓ | AbsSum ↓ |
| Left | CA1 | 0.04 | 0.91 | **0.03** | **0.74** | 0.04 | 0.97 | 0.04 | 0.98 |
| | CA2 | 0.6 | 11.44 | **0.1** | **2.43** | 0.18 | 4.27 | 0.13 | 2.5 |
| | CA3 | 0.2 | 3.86 | **0.1** | **2.34** | 0.21 | 4.8 | 0.13 | 2.91 |
| | DG | 0.1 | 1.58 | **0.04** | 1.12 | **0.04** | 1.05 | 0.05 | **0.93** |
| | SUB | 0.06 | 1.21 | **0.04** | **1.09** | 0.05 | 1.19 | 0.05 | 1.14 |
| Right | CA1 | **0.03** | **0.86** | 0.06 | 1.19 | 0.08 | 1.41 | 0.05 | 1.03 |
| | CA2 | 0.36 | 6.9 | 0.09 | 2.03 | 0.13 | 2.85 | **0.07** | **1.63** |
| | CA3 | 0.14 | 3.43 | **0.08** | **1.93** | 0.14 | 3.38 | 0.09 | 2.13 |
| | DG | 0.07 | 1.54 | **0.03** | **0.77** | **0.03** | 0.84 | 0.04 | 0.82 |
| | SUB | **0.06** | 1.45 | 0.07 | 1.41 | 0.07 | 1.38 | **0.06** | **1.29** |

Comparing the sum of absolute differences between Table 6 and Table 7 along with their standard deviations reveals that our proposed model demonstrated greater stability than both ASHS and the linear upsampling model in calculating biomarkers across most subfields and subregions.

**Table 7**. Cortical subregion median thickness test-retest stability analysis. The proposed INR-based method achieved better stability. The best result in each row is highlighted in blue and the second-best result is underlined.

| Side | Subregion | ASHS (Anisotropic) | | SuperRes-Seg (Nearly isotropic) | | Linear-Seg (Isotropic) | | Proposed (Isotropic) | |
|---|---|---|---|---|---|---|---|---|---|
| | | SD ↓ | AbsSum ↓ | SD ↓ | AbsSum ↓ | SD ↓ | AbsSum ↓ | SD ↓ | AbsSum ↓ |
| Left | ERC | 0.14 | 2.71 | 0.1 | 1.69 | 0.09 | **1.4** | **0.08** | 1.6 |
| | BA35 | 0.18 | 3.37 | 0.11 | 1.71 | **0.09** | 1.88 | **0.09** | **1.62** |
| | BA36 | 0.09 | **1.5** | 0.09 | 1.63 | 0.12 | 1.96 | **0.08** | 1.72 |
| | PHC | 0.11 | 1.91 | **0.05** | **0.9** | 0.08 | 1.62 | 0.06 | 1.03 |
| Right | ERC | 0.18 | 3.1 | 0.08 | 1.35 | **0.07** | **1.22** | 0.09 | 1.7 |
| | BA35 | 0.16 | 3.08 | 0.1 | 1.82 | 0.11 | 1.87 | **0.09** | **1.53** |

| | | | | | | | | |
|---|---|---|---|---|---|---|---|---|
| | BA36 | <u>0.1</u> | <u>1.82</u> | 0.11 | 1.89 | 0.11 | 2.12 | **0.06** | **1.3** |
| | PHC | 0.11 | 1.67 | <u>0.07</u> | <u>1.34</u> | 0.08 | 1.61 | **0.06** | **1.06** |

### *3.6 Pointwise comparison between CU and MCI in template space*

The CRASHS method computed the pointwise thickness of the brain regions, and we compared whether the region thicknesses of the CU and MCI participants differed significantly in a template space. Since these two groups were clinically diagnosed and PET-confirmed, we expected that an accurate subregion segmentation would reflect the true structural changes in these two groups. Figure 6 shows the t-statistic maps calculated based on segmentations from different models.

As shown in Figure 6, the ASHS model shows low-magnitude t-statistics over a larger area of the surface than the other models, which indicates a weaker group difference across that region. Once the segmentation resolution is improved, high-magnitude t-statistics appear over a larger area of the template surface. Our proposed model shows these high-magnitude values over the largest area. Compared with the Linear-Seg method, CA1, CA2, and SUB show a denser red, indicating stronger differences, as pointed out by the light blue arrows in Figure 6. The difference from the SuperRes-Seg method is even clearer, as the SuperRes-Seg surface shows a much lighter red. The pointwise p-values are provided in Supplementary Material S7.

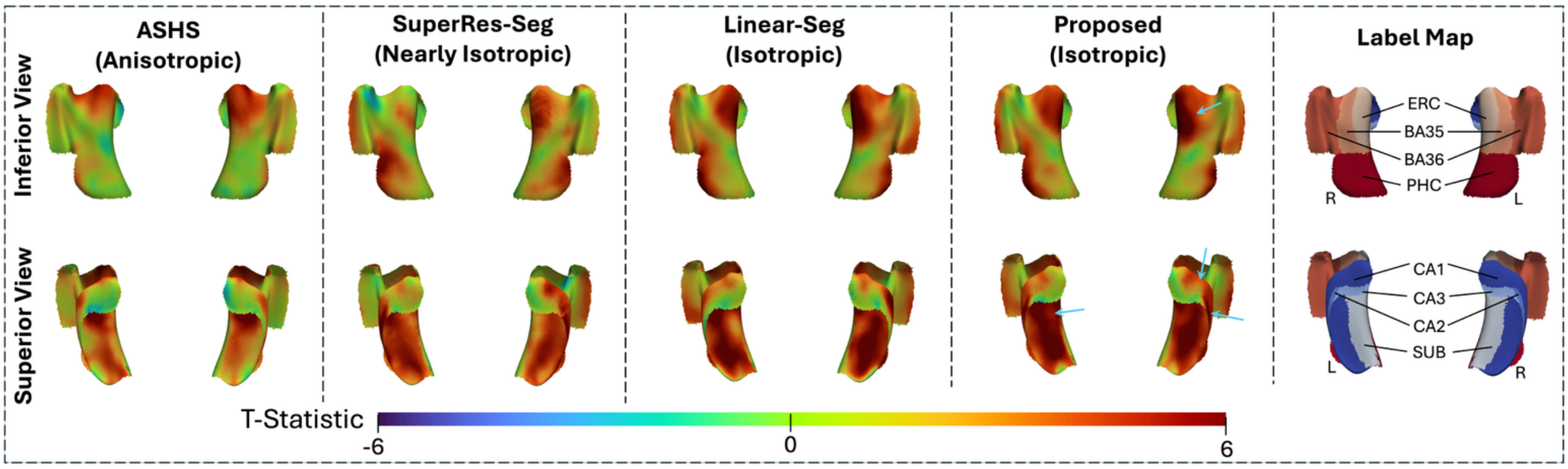

**Figure 6**. Pointwise group difference of thickness between CU and MCI participants.

## 4 Discussion and Conclusions

In this study, we proposed an MTL subregion segmentation framework in an isotropic space. Compared to segmentation in anisotropic space, this method enabled more accurate biomarker extraction for MTL subregions, effectively distinguishing between CU and MCI groups, while also providing more stable measurements over time within the CU group.

Prior to our research, ASHS had already been widely applied as a mature tool for segmentation of MTL subregions. However, since the resolution of ASHS's output is same as that of its input, segmentation result is anisotropic when the input image resolution is anisotropic (e.g., T2w image). In such cases, the boundary of segmentation becomes highly staircase-like, introducing noise into biomarker extraction. Therefore, this study addressed this issue by proposing a high-precision isotropic segmentation method. It also improved the performance of other downstrem pipeline, such as the MTL surface analysis method CRASHS. We summarize its advantages as follows:

One advantage of isotropic segmentation is its impact on volume and thickness calculation. When calculating volume, voxels in MR images may contain information from multiple tissues due to the partial volume effects. At low resolutions, each voxel represents a larger volume, and the inclusion or exclusion of voxels at the boundaries of tissues introduces greater uncertainty in subregion volume calculation. Segmentation in high-resolution spaces can effectively reduce this uncertainty. Furthermore, the structure of the cerebral cortex extends in all directions. To calculate cortical thickness, we need to measure the distance between gray/white and pial surfaces (or, alternatively, between the geometric skeleton and these surfaces). If the resolution is anisotropic, the accuracy of this distance measurement can vary significantly depending on the direction. For instance, distances measured parallel to the coronal view tend to be more accurate since the coronal slice offers the highest in-plane resolution. In contrast, our proposed isotropic segmentation avoids this issue because it maintains the same resolution in all three dimensions, ensuring that direction does not influence the thickness calculation.

Another advantage of this method is that it standardizes the resolution across different participants and scanning sessions. This has two key aspects. First, it aligns the resolution between the training set and the test set, which enhances the model's generalizability. The original resolution of the training set is approximately $0.4 \times 0.4 \times 2.6$ mm$^3$, while the resolution of the test set is about $0.4 \times 0.4 \times 1.2$ mm$^3$. In our preliminary experiment (supplementary materials S3), nnU-Net performed well when the training and test sets had similar resolution distributions, however, its generalization ability was not as strong as that of conventional registration-based methods when there was a significant difference in resolution between the two. By using isotropic upsampling, the training and test sets achieve the same isotropic resolution, allowing the high accuracy of nnU-Net to be fully utilized. Second, within the test set itself, coronal slice thicknesses may vary, such as in the current case in which thickness was 1.2, 1.6, and 2.0 mm. Conducting statistical analysis on a dataset with such varying distributions can lead to biases when analyzing thickness and volume. Isotropic upsampling unifies the slice thickness across test cases, ensuring that the subsequent statistical analyses are conducted within the same and more refined measurement space.

Comparing our proposed method with the other two (nearly) isotropic methods, SuperRes-Seg and Linear-Seg, the results can be examined at two levels. The direct level is the Dice comparison, since the Dice score is the standard way to train and evaluate a segmentation model. Here our proposed model outperformed all the baselines, which shows a clear improvement on the segmentation task itself. The downstream level covers the group discrimination and test-retest tasks, where the measurements are extracted from the segmentation. At this level, the advantage of our model was smaller, and in some subregions the best result came from a baseline. For example, in the test-retest evaluation, SuperRes-Seg achieved almost all the best results in the left hippocampal subfields, while our model was second-best there. This may not mean that the segmentation improvement is lost. A downstream measurement could reduce the sensitivity to the segmentation quality in different ways. For the median thickness, each subregion is compressed into a single value, which is relatively stable, so a more accurate segmentation may not change it much. For the group discrimination, a method that does not give the best discrimination result is not necessarily worse. For example, a segmentation that consistently under-segments each region by a fixed proportion could still distinguish the two groups.

More importantly, our method has clear advantages in the pipeline and resolution. SuperRes-Seg needs two pretrained networks for the upsampling, the SuperRes model that upsamples the manual segmentation by a factor of five along the slice thickness, and the CSF-fixing network that corrects the CSF labels missed during upsampling. Both depend on paired training data and make the pipeline longer. Our model needs no pretrained upsampling step and upsamples each case directly. Its pipeline is as short as linear upsampling, but it produces a more realistic anatomical structure. As for the resolution, our proposed method can produce real isotropic segmenetation while SuperRes-Seg can only be performed at slice thickness of 0.52 mm.

In the discrimination between MCI and CU, our proposed method achieved the highest AUC from the left CA1 volume, at 0.87. For the cortical median thickness, the best AUC was obtained from the left ERC, at 0.722. This pattern, in which the subfield volumes give a higher AUC than the cortical subregions, was also seen in some other studies. For example, (Gertje et al., 2016) reported the highest hippocampal AUC of 0.84 (CA1), higher than the 0.81 (ERC) in the cortical subregions. (Yushkevich et al., 2024) reported the highest hippocampal AUC of 0.77 (DG), higher than the 0.73 (BA35), when distinguishing normal participants from a pathology-defined group that mostly consisted of MCI cases. Across these studies, the AUC values vary over a wide range. For CA1, the region most frequently reported, the AUC was 0.88 in (La Joie et al., 2013), 0.84 in (Pluta et al., 2012), 0.84 in (Gertje et al., 2016), and 0.75 in (Yushkevich et al., 2024). One important reason for this variation is that these studies did not use exactly the same atlas and test dataset. Our proposed model achieved an AUC of 0.87 for CA1, which is at a relatively high level.

For hippocampal subfield segmentation, FreeSurfer and HippUnfold (DeKraker et al., 2022) are also popular methods. Their segmentation results exhibit isotropic resolution. One difference between our method and them is that we take advantage of ASHS's in vivo image atlas, which is more customizable and more suitable for users to modify according to their own needs, such as (Sadeghpour et al., 2025). FreeSurfer's atlas requires not only in vivo T1w images to provide templates for surrounding brain structures, but also ex vivo 7T 3D FLASH T2*-weighted images to provide segmentation templates for hippocampal subfields. This demand for ex vivo data increases the difficulty of atlas acquisition. HippUnfold requires histology images to construct its atlas, which are difficult to obtain. Another difference in our method is that our atlas includes not only hippocampal subfield labels but also complete MTL cortex subregions. This enables biomarker extraction and analysis across more regions.

Lack of evaluation with an external positive control can be a limitation of this study. This is because the initial goal of this study is to address the anisotropy of the segmentation in ASHS, the comparison in the results is within the ASHS family. As a comparison orthogonal to the resolution comparison, two other popular subfield segmentation methods, FreeSurfer and HippUnfold (DeKraker et al., 2022), are also included in the supplementary materials as a reference. One difference between our method and them is that we take advantage of ASHS's in vivo image atlas, which is more customizable and more suitable for users to modify according to their own needs, such as (Sadeghpour et al., 2025). FreeSurfer's atlas requires not only in vivo T1w images to provide templates for surrounding brain structures, but also ex vivo 7T 3D FLASH T2*-weighted images to provide segmentation templates for hippocampal subfields. This demand for ex vivo data increases the difficulty of atlas acquisition. HippUnfold requires histology images to construct its atlas, which are difficult to obtain. Another difference in our method is that our atlas includes not only hippocampal subfield labels but also complete MTL cortex subregions. This enables biomarker extraction and analysis across more regions. In future work, we can incorporate the strengths of FreeSurfer and HippUnfold into our framework. Specifically, we can follow FreeSurfer in further dividing the subregions into fine-grained subregions to provide more detail. We can also follow HippUnfold in using histology images to define the subregion borders based on cytoarchitecture.

The test-retest analysis is based on the assumption that a more stable model produces relatively consistent measurements across repeated scans. However, the consistency may be affected by several other factors. One important factor is that we used longitudinal data for the test-retest analysis. Even for CU participants, normal ageing can still cause structural changes, so some change in the measurements is unavoidable, and it may introduce variability into the evaluation. In addition, other factors, such as the imaging quality and the noise in the measurement calculation, like the skeleton and radius calculations, may also cause changes in the measurements. These changes can affect the test-retest evaluation, but they do not come from the segmentation itself. This is a limitation of using longitudinal data on test-retest analysis.

A limitation of this model is that it does not use the upsampled image from INR. Although INR can generate the upsampled T1w and T2w images, it is trained case by case, which means we have to train an individual INR for each test case. Such a process requires about half an hour for each side of the MTL ROI, which is orders of magnitude longer than the inference time for the propsoed segmentation model (~38 s). Therefore, the image upsampling relies on conventional linear interpolation to keep the inference fast. In the future, we can explore using a generative methods like diffusion models as an alternative or complement to INR super-resolution. It should accelerate the image upsampling speed and ensure the segmentation model takes high-quality images as input.

A second limitation is that the isotropic resolution used in this study is only defined according to the in-plane resolution of the coronal slice and is not the upper bound of isotropic resolution. As suggested in (Li et al., 2025a), isotropic high-resolution ex vivo MRI ($0.2 \times 0.2 \times 0.2$ mm$^3$) can be used to train an upsampling model for both image and segmentation. In the future, we will combine the INR model and ex vivo image-guided model to design a super-resolution segmentation method to make full use of the resolution and extract more accurate imaging biomarkers.

The isotropic multi-modality MTL subregion segmentation model proposed in this study fully utilizes the structural, contrast, and intensity advantages of different modalities, as well as their resolution advantages, to achieve more accurate and stable calculation of subregion thickness. This model helps improve the accuracy of early AD imaging biomarkers without increasing the cost of image acquisition.

# Highlights

- Anisotropic resolution hinders the Alzheimer's disease imaging biomarker extraction.
- An isotropic MTL subregion segmentation training dataset was constructed.
- A multi-modality MTL subregion segmentation pipeline was established.
- Isotropic segmentation enhanced the accuracy and stability of imaging biomarkers.

# Ethics Statement

The test data was collected from Penn Alzheimer’s Disease Research Center (ADRC) cohort, where written informed consent was obtained from all Penn ADRC participants according to the Declaration of Helsinki under a protocol approved by the University of Pennsylvania Institutional Review Board.

# Acknowledgements

This work was supported by NIH (grant numbers R01-AG069474, RF1-AG056014, P30-AG072979, R01-AG070592). LW is supported by MultiPark, a strategic research area at Lund University, and project grants from the Bente Rexed Gerstedt Foundation, Fondation Recherche Alzheimer and the Crafoord Foundation (20230790).

## Data and Code Availability

The upsampled atlas is available at https://doi.org/10.5061/dryad.k6djh9wmn (under review for now) and the code is available at https://github.com/liyue3780/HyperResASHS. The trained model is available at https://huggingface.co/upennpatchlab/hyperresashs_model_pmc_t2_v1/tree/main.

## References

Backhausen, L.L., Herting, M.M., Tamnes, C.K., Vetter, N.C., 2022. Best Practices in Structural Neuroimaging of Neurodevelopmental Disorders. Neuropsychol Rev 32, 400–418. https://doi.org/10.1007/s11065-021-09496-2

Benzinger, T.L.S., Blazey, T., Jack, C.R., Koeppe, R.A., Su, Y., Xiong, C., Raichle, M.E., Snyder, A.Z., Ances, B.M., Bateman, R.J., Cairns, N.J., Fagan, A.M., Goate, A., Marcus, D.S., Aisen, P.S., Christensen, J.J., Ercole, L., Hornbeck, R.C., Farrar, A.M., Aldea, P., Jasielec, M.S., Owen, C.J., Xie, X., Mayeux, R., Brickman, A., McDade, E., Klunk, W., Mathis, C.A., Ringman, J., Thompson, P.M., Ghetti, B., Saykin, A.J., Sperling, R.A., Johnson, K.A., Salloway, S., Correia, S., Schofield, P.R., Masters, C.L., Rowe, C., Villemagne, V.L., Martins, R., Ourselin, S., Rossor, M.N., Fox, N.C., Cash, D.M., Weiner, M.W., Holtzman, D.M., Buckles, V.D., Moulder, K., Morris, J.C., 2013. Regional variability of imaging biomarkers in autosomal dominant Alzheimer's disease. Proc. Natl. Acad. Sci. U.S.A. 110. https://doi.org/10.1073/pnas.1317918110

Braak, H., Braak, E., 1995. Staging of alzheimer's disease-related neurofibrillary changes. Neurobiology of Aging 16, 271–278. https://doi.org/10.1016/0197-4580(95)00021-6

Chen, Y., Liu, S., Wang, X., 2021. Learning Continuous Image Representation with Local Implicit Image Function, in: 2021 IEEE/CVF Conference on Computer Vision and Pattern Recognition (CVPR). Presented at the 2021 IEEE/CVF Conference on Computer Vision and Pattern Recognition (CVPR), IEEE, Nashville, TN, USA, pp. 8624–8634. https://doi.org/10.1109/CVPR46437.2021.00852

DeKraker, J., Cabalo, D., Royer, J., Khan, A., Karat, B., Benkarim, O., Cruces, R.R., Frauscher, B., Pana, R., Hansen, J., Misic, B., Valk, S., Lau, J., Kirschner, M., Bernsconi, A., Bernasconi, N., Muenzing, S., Axer, M., Amunts, K., Evans, A., Bernhardt, B., 2024. HippoMaps: Multiscale Cartography of Human Hippocampal Organization (preprint). In Review. https://doi.org/10.21203/rs.3.rs-4045651/v1

DeKraker, J., Haast, R.A., Yousif, M.D., Karat, B., Lau, J.C., Köhler, S., Khan, A.R., 2022. Automated hippocampal unfolding for morphometry and subfield segmentation with HippUnfold. eLife 11, e77945. https://doi.org/10.7554/eLife.77945

Fischl, B., 2012. FreeSurfer. NeuroImage 62, 774–781. https://doi.org/10.1016/j.neuroimage.2012.01.021

Fjell, A.M., Walhovd, K.B., Fennema-Notestine, C., McEvoy, L.K., Hagler, D.J., Holland, D., Brewer, J.B., Dale, A.M., 2009. One-Year Brain Atrophy Evident in Healthy Aging. J. Neurosci. 29, 15223–15231. https://doi.org/10.1523/JNEUROSCI.3252-09.2009

Gertje, E.C., Pluta, J., Das, S., Mancuso, L., Kliot, D., Yushkevich, P., Wolk, D., 2016. Clinical Application of Automatic Segmentation of Medial Temporal Lobe Subregions in Prodromal and Dementia-Level Alzheimer's Disease. JAD 54, 1027–1037. https://doi.org/10.3233/JAD-160014

Hampel, H., Bürger, K., Teipel, S.J., Bokde, A.L.W., Zetterberg, H., Blennow, K., 2008. Core candidate neurochemical and imaging biomarkers of Alzheimer's disease⁎. Alzheimer's & Dementia 4, 38–48. https://doi.org/10.1016/j.jalz.2007.08.006

Iglesias, J.E., Augustinack, J.C., Nguyen, K., Player, C.M., Player, A., Wright, M., Roy, N., Frosch, M.P., McKee, A.C., Wald, L.L., Fischl, B., Van Leemput, K., 2015. A computational atlas of the hippocampal formation using ex vivo, ultra-high resolution MRI: Application to adaptive segmentation of in vivo MRI. NeuroImage 115, 117–137. https://doi.org/10.1016/j.neuroimage.2015.04.042

Isensee, F., Jaeger, P.F., Kohl, S.A.A., Petersen, J., Maier-Hein, K.H., 2021. nnU-Net: a self-configuring method for deep learning-based biomedical image segmentation. Nat Methods 18, 203–211. https://doi.org/10.1038/s41592-020-01008-z

Jack, C.R., Andrews, J.S., Beach, T.G., Buracchio, T., Dunn, B., Graf, A., Hansson, O., Ho, C., Jagust, W., McDade, E., Molinuevo, J.L., Okonkwo, O.C., Pani, L., Rafii, M.S., Scheltens, P., Siemers, E., Snyder, H.M., Sperling, R., Teunissen, C.E., Carrillo, M.C., 2024. Revised criteria for diagnosis and staging of Alzheimer's disease: Alzheimer's Association Workgroup. Alzheimer's & Dementia 20, 5143–5169. https://doi.org/10.1002/alz.13859

Kreilkamp, B.A.K., Weber, B., Elkommos, S.B., Richardson, M.P., Keller, S.S., 2018. Hippocampal subfield segmentation in temporal lobe epilepsy: Relation to outcomes. Acta Neurol Scand 137, 598–608. https://doi.org/10.1111/ane.12926

La Joie, R., Perrotin, A., de La Sayette, V., Egret, S., Doeuvre, L., Belliard, S., Eustache, F., Desgranges, B., Chételat, G., 2013. Hippocampal subfield volumetry in mild cognitive impairment, Alzheimer's disease and semantic dementia. NeuroImage: Clinical 3, 155–162. https://doi.org/10.1016/j.nicl.2013.08.007

Li, Y., Khandelwal, P., Xie, L., Wisse, L.E.M., Mundada, N., Brown, C.A., McGrew, E., Denning, A., Das, S.R., Wolk, D.A., Yushkevich, P.A., 2025a. Nearly isotropic segmentation for medial temporal lobe subregions in multi-modality MRI. https://doi.org/10.48550/arXiv.2504.18442

Li, Y., Xie, L., Khandelwal, P., Wisse, L.E.M., Brown, C.A., Prabhakaran, K., Tisdall, M.D., Mechanic-Hamilton, D., Detre, J.A., Das, S.R., Wolk, D.A., Yushkevich, P.A., 2025b. Automatic Segmentation of Medial Temporal Lobe Subregions in Multi-Scanner, Multi-Modality Magnetic Resonance Imaging of Variable Quality. Hippocampus 35, e70036. https://doi.org/10.1002/hipo.70036

Manjón, J.V., Coupé, P., Buades, A., Fonov, V., Louis Collins, D., Robles, M., 2010. Non-local MRI upsampling. Medical Image Analysis 14, 784–792. https://doi.org/10.1016/j.media.2010.05.010

Márquez, F., Yassa, M.A., 2019. Neuroimaging Biomarkers for Alzheimer's Disease. Mol Neurodegeneration 14, 21. https://doi.org/10.1186/s13024-019-0325-5

McGinnis, J., Shit, S., Li, H.B., Sideri-Lampretsa, V., Graf, R., Dannecker, M., Pan, J., Stolt-Ansó, N., Mühlau, M., Kirschke, J.S., Rueckert, D., Wiestler, B., 2023. Single-subject Multi-contrast MRI Super-resolution via Implicit Neural Representations, in: Greenspan, H., Madabhushi, A., Mousavi, P., Salcudean, S., Duncan, J., Syeda-Mahmood, T., Taylor, R. (Eds.), Medical Image Computing and Computer Assisted Intervention – MICCAI 2023, Lecture Notes in Computer Science. Springer Nature Switzerland, Cham, pp. 173–183. https://doi.org/10.1007/978-3-031-43993-3_17

Mildenhall, B., Srinivasan, P.P., Tancik, M., Barron, J.T., Ramamoorthi, R., Ng, R., 2022. NeRF: representing scenes as neural radiance fields for view synthesis. Commun. ACM 65, 99–106. https://doi.org/10.1145/3503250

Mintun, M.A., Lo, A.C., Duggan Evans, C., Wessels, A.M., Ardayfio, P.A., Andersen, S.W., Shcherbinin, S., Sparks, J., Sims, J.R., Brys, M., Apostolova, L.G., Salloway, S.P., Skovronsky, D.M., 2021. Donanemab in Early Alzheimer's Disease. N Engl J Med 384, 1691–1704. https://doi.org/10.1056/NEJMoa2100708

Mueller, S.G., Yushkevich, P.A., Das, S., Wang, L., Van Leemput, K., Iglesias, J.E., Alpert, K., Mezher, A., Ng, P., Paz, K., Weiner, M.W., 2018. Systematic comparison of different techniques to measure hippocampal subfield volumes in ADNI2. NeuroImage: Clinical 17, 1006–1018. https://doi.org/10.1016/j.nicl.2017.12.036

Nelson, P.T., Alafuzoff, I., Bigio, E.H., Bouras, C., Braak, H., Cairns, N.J., Castellani, R.J., Crain, B.J., Davies, P., Tredici, K.D., Duyckaerts, C., Frosch, M.P., Haroutunian, V., Hof, P.R., Hulette, C.M., Hyman, B.T., Iwatsubo, T., Jellinger, K.A., Jicha, G.A., Kövari, E., Kukull, W.A., Leverenz, J.B., Love, S., Mackenzie, I.R., Mann, D.M., Masliah, E., McKee, A.C., Montine, T.J., Morris, J.C., Schneider, J.A., Sonnen, J.A., Thal, D.R., Trojanowski, J.Q., Troncoso, J.C., Wisniewski, T., Woltjer, R.L., Beach, T.G., 2012. Correlation of Alzheimer Disease Neuropathologic Changes With Cognitive Status: A Review of the Literature. J Neuropathol Exp Neurol 71, 362–381. https://doi.org/10.1097/NEN.0b013e31825018f7

Ogniewicz, R.L., Kübler, O., 1995. Hierarchic Voronoi Skeletons. Pattern Recognition 28, 343–359. https://doi.org/doi.org/10.1016/0031-3203(94)00105-U

Pluta, J., Yushkevich, P., Das, S., Wolk, D., 2012. In vivo Analysis of Hippocampal Subfield Atrophy in Mild Cognitive Impairment via Semi-Automatic Segmentation of T2-Weighted MRI. JAD 31, 85–99. https://doi.org/10.3233/JAD-2012-111931

Pouch, A.M., Tian, S., Takebe, M., Yuan, J., Gorman, R., Cheung, A.T., Wang, H., Jackson, B.M., Gorman, J.H., Gorman, R.C., Yushkevich, P.A., 2015. Medially constrained deformable modeling for segmentation of branching medial structures: Application to aortic valve segmentation and morphometry. Medical Image Analysis 26, 217–231. https://doi.org/10.1016/j.media.2015.09.003

Romero, J.E., Coupé, P., Manjón, J.V., 2017. HIPS: A new hippocampus subfield segmentation method. NeuroImage 163, 286–295. https://doi.org/10.1016/j.neuroimage.2017.09.049

Sadeghpour, N., Lim, S.A., Wuestefeld, A., Denning, A.E., Ittyerah, R., Trotman, W., Chung, E., Sadaghiani, S., Prabhakaran, K., Bedard, M.L., Ohm, D.T., Artacho-Pérula, E., de Onzoño Martin, M.M.I., Muñoz, M., Romero, F.J.M., González, J.C.D., del Arroyo Jiménez, M., del Marcos Rabal, M., Insausti Serrano, A.M., González, N.V., Sánchez, S.C., de la Rosa Prieto, C., Insausti, R., McMillan, C., Lee, E.B., Detre, J.A., Das, S.R., Xie, L., Tisdall, M.D., Irwin, D.J., Wolk, D.A., Yushkevich, P.A., Wisse, L.E.M., 2025. Developing an Anatomically Valid Segmentation Protocol for Anterior Regions of the Medial Temporal Lobe for Neurodegenerative Diseases. Hippocampus 35, e70027. https://doi.org/10.1002/hipo.70027

Scheltens, P., Blennow, K., Breteler, M.M.B., Strooper, B. de, Frisoni, G.B., Salloway, S., Flier, W.M.V. der, 2016. Alzheimer's disease. The Lancet 388, 505–517. https://doi.org/10.1016/S0140-6736(15)01124-1

Sitzmann, V., Martel, J., Bergman, A., Lindell, D., Wetzstein, G., 2020. Implicit Neural Representations with Periodic Activation Functions, in: Advances in Neural Information Processing Systems. Curran Associates, Inc., pp. 7462–7473.

Sone, D., Sato, N., Maikusa, N., Ota, M., Sumida, K., Yokoyama, K., Kimura, Y., Imabayashi, E., Watanabe, Y., Watanabe, M., Okazaki, M., Onuma, T., Matsuda, H., 2016. Automated subfield volumetric analysis of hippocampus in temporal lobe epilepsy using high-resolution T2-weighed MR imaging. NeuroImage: Clinical 12, 57–64. https://doi.org/10.1016/j.nicl.2016.06.008

Tancik, M., Srinivasan, P., Mildenhall, B., Fridovich-Keil, S., Raghavan, N., Singhal, U., Ramamoorthi, R., Barron, J., Ng, R., 2020. Fourier Features Let Networks Learn High Frequency Functions in Low Dimensional Domains, in: Advances in Neural Information Processing Systems. Curran Associates, Inc., pp. 7537–7547.

van Dyck Christopher H., Swanson Chad J., Aisen Paul, Bateman Randall J., Chen Christopher, Gee Michelle, Kanekiyo Michio, Li David, Reyderman Larisa, Cohen Sharon, Froelich Lutz, Katayama Sadao, Sabbagh Marwan, Vellas Bruno, Watson David, Dhadda Shobha, Irizarry Michael, Kramer Lynn D., Iwatsubo Takeshi, 2023. Lecanemab in Early Alzheimer's Disease. New England Journal of Medicine 388, 9–21. https://doi.org/10.1056/NEJMoa2212948

Venet, L., Pati, S., Feldman, M.D., Nasrallah, M.P., Yushkevich, P., Bakas, S., 2021. Accurate and Robust Alignment of Differently Stained Histologic Images Based on Greedy Diffeomorphic Registration. Applied Sciences 11, 1892. https://doi.org/10.3390/app11041892

Ver Hoef, L., Deshpande, H., Cure, J., Selladurai, G., Beattie, J., Kennedy, R.E., Knowlton, R.C., Szaflarski, J.P., 2021. Clear and Consistent Imaging of Hippocampal Internal Architecture With High Resolution Multiple Image Co-registration and Averaging (HR-MICRA). Front. Neurosci. 15, 546312. https://doi.org/10.3389/fnins.2021.546312

Wisse, L.E.M., Adler, D.H., Ittyerah, R., Pluta, J.B., Robinson, J.L., Schuck, T., Trojanowski, J.Q., Grossman, M., Detre, J.A., Elliott, M.A., Toledo, J.B., Liu, W., Pickup, S., Das, S.R., Wolk, D.A., Yushkevich, P.A., 2016. Comparison of In Vivo and Ex Vivo MRI of the Human Hippocampal Formation in the Same Subjects. Cereb. Cortex cercor;bhw299v1. https://doi.org/10.1093/cercor/bhw299

Wisse, L.E.M., Daugherty, A.M., Olsen, R.K., Berron, D., Carr, V.A., Stark, C.E.L., Amaral, R.S.C., Amunts, K., Augustinack, J.C., Bender, A.R., Bernstein, J.D., Boccardi, M., Bocchetta, M., Burggren, A., Chakravarty, M.M., Chupin, M., Ekstrom, A., de Flores, R., Insausti, R., Kanel, P., Kedo, O., Kennedy, K.M., Kerchner, G.A., LaRocque, K.F., Liu, X., Maass, A., Malykhin, N., Mueller, S.G., Ofen, N., Palombo, D.J., Parekh, M.B., Pluta, J.B., Pruessner, J.C., Raz, N., Rodrigue, K.M., Schoemaker, D., Shafer, A.T., Steve, T.A., Suthana, N., Wang, L., Winterburn, J.L., Yassa, M.A., Yushkevich, P.A., la Joie, R., Group, for the H.S., 2017. A harmonized segmentation protocol for hippocampal and parahippocampal subregions: Why do we need one and what are the key goals? Hippocampus 27, 3–11. https://doi.org/10.1002/hipo.22671

Xie, L., Wisse, L.E.M., Pluta, J., De Flores, R., Piskin, V., Manjón, J.V., Wang, H., Das, S.R., Ding, S., Wolk, D.A., Yushkevich, P.A., for the Alzheimer's Disease Neuroimaging Initiative, 2019. Automated segmentation of medial temporal lobe subregions on in vivo T1-weighted MRI in early stages of Alzheimer's disease. Human Brain Mapping 40, 3431–3451. https://doi.org/10.1002/hbm.24607

Xie, L., Wisse, L.E.M., Wang, J., Ravikumar, S., Khandelwal, P., Glenn, T., Luther, A., Lim, S., Wolk, D.A., Yushkevich, P.A., 2023. Deep label fusion: A generalizable hybrid multi-atlas and deep convolutional neural network for medical image segmentation. Medical Image Analysis 83, 102683. https://doi.org/10.1016/j.media.2022.102683

Yushkevich, P.A., Ittyerah, R., Li, Y., Denning, A.E., Sadeghpour, N., Lim, S., McGrew, E., Xie, L., DeFlores, R., Brown, C.A., Wisse, L.E.M., Wolk, D.A., Das, S.R., for the Alzheimer's Disease Neuroimaging Initiative, 2024. Morphometry of medial temporal lobe subregions using high-resolution T2-weighted MRI in ADNI3: Why, how, and what's next? Alzheimer's & Dementia 20, 8113–8128. https://doi.org/10.1002/alz.14161

Yushkevich, P.A., Pluta, J.B., Wang, H., Xie, L., Ding, S.-L., Gertje, E.C., Mancuso, L., Kliot, D., Das, S.R., Wolk, D.A., 2015. Automated volumetry and regional thickness analysis of hippocampal subfields and medial temporal cortical structures in mild cognitive impairment: Automatic Morphometry of MTL Subfields in MCI. Hum. Brain Mapp. 36, 258–287. https://doi.org/10.1002/hbm.22627